\documentclass[preprint,12pt]{elsarticle}
\usepackage{amssymb}
\usepackage{amsmath,graphicx,algorithm,algorithmicx,subfigure,multicol,multirow,amssymb,bm}
\usepackage{array}
\usepackage{booktabs}
\usepackage{algpseudocode}
\usepackage{caption}
\usepackage{tabularx}
\usepackage{threeparttable}
\usepackage{dcolumn}
\usepackage{longtable}
\usepackage{color}
\usepackage{url}
\newcolumntype{.}{D{.}{.}{-1}}

\makeatletter
\def\hlinew#1{%
  \noalign{\ifnum0=`}\fi\hrule \@height #1 \futurelet
   \reserved@a\@xhline}
\makeatother

\journal{Pattern Recognition}

\begin{document}

\begin{frontmatter}

\title{Discriminative Representation Combinations for Accurate Face Spoofing Detection}

\author{Xiao~Song,~Xu~Zhao,~Liangji~Fang
        and~Tianwei~Lin}
\address{Key Laboratory of System Control and Information Processing MOE,\\ Department of Automation, Shanghai Jiao Tong University, China}

\begin{abstract}
Three discriminative representations for face presentation attack detection are introduced in this paper. Firstly we design a descriptor called spatial pyramid coding micro-texture (SPMT) feature to characterize local appearance information. Secondly we utilize the SSD, which is a deep learning framework for detection, to excavate context cues and conduct end-to-end face presentation attack detection. Finally we design a descriptor called template face matched binocular depth (TFBD) feature to characterize stereo structures of real and fake faces. For accurate presentation attack detection, we also design two kinds of representation combinations. Firstly, we propose a decision-level cascade strategy to combine SPMT with SSD. Secondly, we use a simple score fusion strategy to combine face structure cues (TFBD) with local micro-texture features (SPMT). To demonstrate the effectiveness of our design, we evaluate the representation combination of SPMT and SSD on three public datasets, which outperforms all other state-of-the-art methods. In addition, we evaluate the representation combination of SPMT and TFBD on our dataset and excellent performance is also achieved.

\end{abstract}


\end{frontmatter}

\clearpage

\section{Introduction}
\label{s1}

In recent years, face recognition based identity authentication systems \cite{deng2017fine,zhou2018age} are popular.
However, similar to other biometric modalities \cite{rattani2012analysis,evans2013spoofing}, security risks hide in the system. Many authentication systems can't judge whether faces are captured from authorized clients or from presentation attacks.

\begin{figure}[tb]
\captionsetup{font={footnotesize}}
\centering
\includegraphics[width=4.2in,height=2.3in]{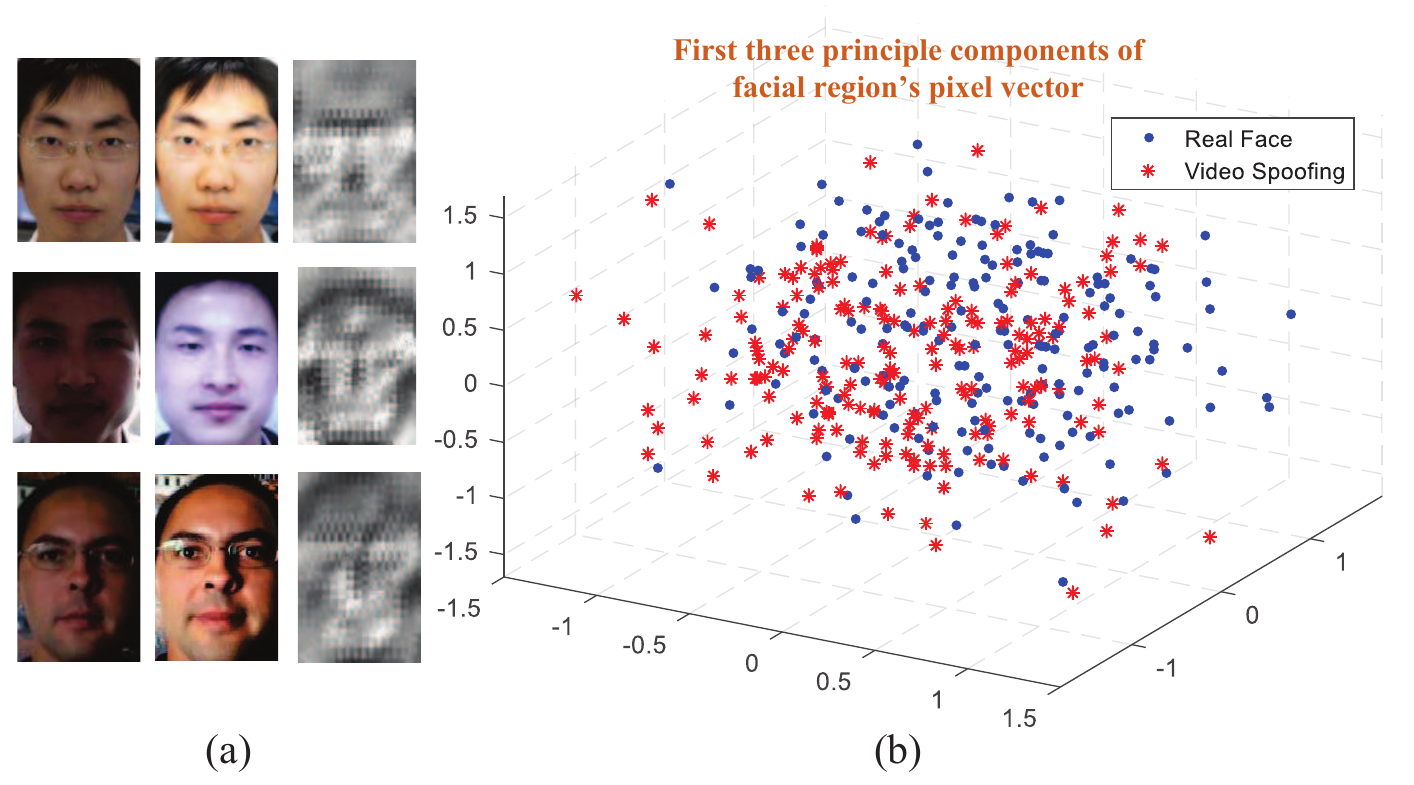}
\caption{(a) From top to bottom, NUAA, CASIA and REPLAY-ATTACK datasets, from left to right, real face, fake face and Fisher face. (b) We randomly select $200$ real faces and $200$ video attack frames from REPLAY-ATTACK. Pixel vectors in facial regions are hard to distinguish.}
\label{f1}
\end{figure}

There are various presentation attacks, for example, prints, photographs, videos displayed on screens and 3D models such as face masks \cite{liu20163d}. Images or videos of an authorized user can be easily obtained from Internet or by portable cameras. 2D fake faces are cheap to make, but 3D masks are expensive to build and are rare in real applications. Hence in this paper, we focus on 2D presentation attacks including prints, photos and videos. As shown in Fig. \ref{f1}, telling real faces is difficult even for humans. Consequently, robust presentation attack detection (PAD) methods are needed.

Recently, several state-of-the-art face PAD methods are proposed. Wen \emph{et al}. \cite{wen2015face} utilize image distortion analysis for presentation attack detection. Boulkenafet \emph{et al}. \cite{boulkenafet2016face} regard micro-texture in color space as the vital cue for presentation attack detection. Yang \emph{et al}. \cite{yang2015person} design a person-specific model for presentation attack detection and Patel \emph{et al}. \cite{patel2016cross} defend presentation attacks based on convolutional neural networks. More related works are illustrated in Section \ref{s2}.

Recaptured images lose some high-frequency information \cite{tan2010face,li2004live,galbally2014image} because of limited resolution and gaussian blurring. Appearance is also changed due to the abnormal shading on re-imaged surfaces. Furthermore, printing artifacts or noise signatures in captured videos \cite{maatta2012face,de2014face} also exist. Consequently, micro-texture is a helpful cue for discriminating the appearance of real face and fake face.

Another helpful cue for presentation attack detection is the stereo structure of face. A fake face displayed on a screen cannot mimic the structure of a real face \cite{wang2013facex} because the screen is always planar. In addition, a printed fake face cannot mimic the rigid structure of a real face by any operations. Hence recovering face stereo structure is beneficial for face liveness judgment.

Recently convolutional neural network (CNN) based methods achieve excellent performance in many computer vision tasks, such as object detection \cite{liu2016ssd}. Deep representations extracted from convolutional neural network are with rich semantical information. Hence utilizing the deep network for face presentation attack detection is appropriate.

In this paper, we propose two kinds of representation combinations for face presentation attack detection. We demonstrate that both of them achieve excellent performance for face PAD task. The first one combines SPMT with SSD, because local appearance descriptors and global context cues are proven complementary. The second one combines SPMT with TFBD, in which stereo structure cues are exploited and combined with 2D appearance features. Based on our design, two effective solutions for face PAD task is obtained.

The first representation combination only needs a single image. Firstly, we design a hand-crafted descriptor called SPMT to encode local micro-textures. We propose a spatial pyramid encoding algorithm hence SPMT is capable of encoding multi-scale information in facial regions. Secondly, we utilize the Single Shot MultiBox Detector (SSD) \cite{liu2016ssd} for end-to-end liveness judgment. As current PAD datasets are relatively small, we reduce the complexity of prediction layers to avoid over-fitting problem. Lastly, we propose a decision-level cascade strategy to confirm or correct obscure judgments from SSD. Once the output is regarded uncertain, SPMT descriptor is extracted from corresponding facial region to make further judgment.

The second representation combination needs a binocular image pair.
We design another hand-crafted descriptor called TFBD to capture stereo structure's difference between real face and fake face. As recovering dense 3D structure is time-consuming, we choose to recover the sparse face structure based on sparse facial landmarks using proposed template face registration algorithm. Finally, TFBD descriptor is combined with SPMT descriptor. The score fusion of two SVM outputs determines the classification result.

In this paper, our main contribution is three-fold.

(\expandafter{\romannumeral1}) We propose the TFBD descriptor with a template face registration algorithm, and the SPMT descriptor with a spatial pyramid encoding algorithm. In addition, we are the first to utilize SSD for face PAD.

(\expandafter{\romannumeral2}) We introduce a complementary representation combination ``SPMT + SSD'' for face PAD, and a decision-level cascade strategy. ``SPMT + SSD'' achieves excellent performance on three public datasets.

(\expandafter{\romannumeral3}) We also introduce another representation combination ``SPMT + TFBD'' for face PAD, where 2D appearance descriptor is proven to be complementary with stereo structure cues.

The rest of this paper is organized as follows. Related work on face presentation attack detection is reviewed in Section \ref{s2}.
The combination of SPMT and SSD is introduced in Section \ref{s3}. The combination of TFBD and SPMT is introduced in Section \ref{s4}. Datasets, protocols, evaluation metrics and experimental results are detailed in Section \ref{s5}. Conclusion and future work are illustrated in Section \ref{s6}.

\section{Related Work}
\label{s2}
Despite the multimodal methods \cite{wild2016robust,huang2013robust}, most face presentation attack detection methods can be divided into five categories: motion based approaches \cite{szwoch2012eye,chetty2010biometric,bao2009liveness,akhtar2015biometric}, texture based approaches \cite{maatta2012face,de2013can,boulkenafet2016face}, stereo structure based approaches \cite{de2012moving,wang2013facex}, deep learning based approaches \cite{yang2014learn,patel2016cross,menotti2015deep} and other approaches \cite{pinto2015face,wen2015face,yan2012face,li2016generalized,bharadwaj2013computationally}. Research surveys can be found in \cite{galbally2014biometric,de2013can}.

(\expandafter{\romannumeral1}) Motion based methods aim at capturing biometric motions such as eye blinking \cite{pan2007eyeblink,szwoch2012eye,sun2007blinking}, mouth movement \cite{kollreider2007real} and holistic facial motions \cite{kollreider2009non,bao2009liveness}. Given a video clip, Pan \emph{et al}. \cite{pan2007eyeblink} regard eye-blink detection as a state transition problem then conditional graphical model is used to model different stages.
In \cite{kollreider2007real}, lip movement and lip-reading are treated as critical cues for presentation attack detection.
Bao \emph{et al}. \cite{bao2009liveness} distinguish planar attack with real face based on motion correlation from optical flow field. Kollreider \emph{et al}. \cite{kollreider2009non} employ an optical flow based model and a local Gabor decomposition model for face motion estimation.
However, these challenge-response approaches require clients' cooperations and the motion cues for presentation attack detection can be easily inferred.

(\expandafter{\romannumeral2}) It is demonstrated in \cite{yang2013face} that local micro-texture is an useful cue for detecting presentation attacks from re-captured images or videos. Maatta \emph{et al}. \cite{maatta2011face} present a novel micro-texture descriptor called Multi-Scale Local Binary Patterns (MSLBP) for face presentation attack detection.
Freitas \emph{et al}.$\;$\cite{de2012lbp} fuse space with time information into a single descriptor called Local Binary Patterns from Three Orthogonal Planes (LBPTOP). Recently, person-specific methods \cite{chingovska2015use,yang2015person} are proposed to improve the generalization ability of micro-texture based algorithms. Zhang \emph{et al}. \cite{zhang2018face} apply the Markov model on color texture features then conduct recursive feature elimination for face PAD. Boulkenafet \emph{et al}. \cite{boulkenafet2016face} focus on luminance and chrominance channels where the joint information of color and texture is exploited. In \cite{boulkenafet2016face}, the same authors propose a solution based on describing the facial appearance by applying Fisher vector encoding on speeded-up robust features. However micro-texture descriptor is low-level thus they are sensitive to illumination changes and images with high quality.

(\expandafter{\romannumeral3}) Stereo structure based methods can be divided into two types: methods without extra hardware and methods requiring extra hardware. For the first type, Maria \emph{et al}. \cite{de2012moving} exploit geometric invariants from a set of automatically located facial landmarks to estimate face structures. Given an image sequence, Yang \emph{et al}. \cite{wang2013face} recover the sparse 3D structure from several selected frames.
Another type utilizes the depth information from depth sensors such as Microsoft Kinect, to reconstruct face structures \cite{erdogmus2013spoofing}. Wang \emph{et al}. \cite{wang2017robust} combine depth information from Kinect with texture features learned from convolutional neural network. However it's difficult to deploy these presentation attack detection systems in real applications. In addition stereo cues may be ineffective when confronting 3D mask attacks.

(\expandafter{\romannumeral4})
Rather than designing hand-crafted features for presentation attack detection, Menotti \emph{et al}. \cite{menotti2015deep} build a robust PAD system for iris, face, and fingerprint modalities based on convolutional neural networks with limited biometric knowledge. Yang \emph{et al}. \cite{yang2014learn} also utilize CNN models to learn deep representations for face PAD task. Gustavo \emph{et al}. \cite{de2017deep} design the LBPnet, in which LBP descriptor is integrated in the first layer of a convolutional neural network then high-level texture features are extracted. Similarly we also incorporate local descriptors with deep features for face PAD task. In addition, to alleviate the problem of over-fitting, Rehman \emph{et al} \cite{rehman2018livenet} proposes a data randomization technique to train CNN classifiers on small-scale face PAD datasets. Moreover, a cross-dataset face PAD algorithm \cite{patel2016cross} is proposed based on a two-stream CaffeNet, however its performance is not good.

(\expandafter{\romannumeral5})
Context cues are also useful for face PAD. Komulainen \emph{et al}. \cite{komulainen2013context} conduct face presentation attack detection by detecting the presentation attack medium in the scene. Yan \emph{et al}. \cite{yan2012face} fuse multiple context cues such as background consistency and scene shift.  However, systems based on these simple cues can be easily cheated hence they are unpractical.

Methods based on image quality analysis are also popular. Galbally \emph{et al}. \cite{galbally2014image} propose an method where image quality metrics are obtained by evaluating prominent factors among $25$ image quality measures. Inspired by \cite{galbally2014image}, Wen \emph{et al}. \cite{wen2015face} extract various image quality representations (specular reflection, blurriness, color diversity) for image distortion analysis and extracted IDA descriptors characterize the inter-class difference. However, these methods are not robust and relatively slow.

Some methods combine different cues mentioned above for PAD, including our previous work \cite{sx2017face}. Pan \emph{et al}. \cite{pan2011monocular} argue that the shift of background scene and eye blinking are both important. Tronci \emph{et al}. \cite{tronci2011fusion} utilize the joint information of motion, texture and image quality to perform both sequential and static liveness analysis. Feng \emph{et al}. \cite{feng2016integration} propose a multi-cues integration framework based on a hierarchical neural network to fuse image quality cues with motion cues.

In this paper, we propose the representation combination of SPMT and SSD to demonstrate the complementarity between local appearance descriptors and context cues from convolutional neural network. In addition, we propose the representation combination of SPMT and TFBD to demonstrate the complementarity between appearance features and stereo structure cues. Compared with our previous work \cite{sx2017face}, proposed descriptors SPMT and TFBD are modified and refined. In addition, we are the first to utilize SSD for face presentation attack detection, and we also propose a strategy to confirm or correct the obscure liveness judgments.

\begin{figure*}[tb]
\captionsetup{font={footnotesize}}
\makebox[\textwidth][c]{\includegraphics[width=1\textwidth]{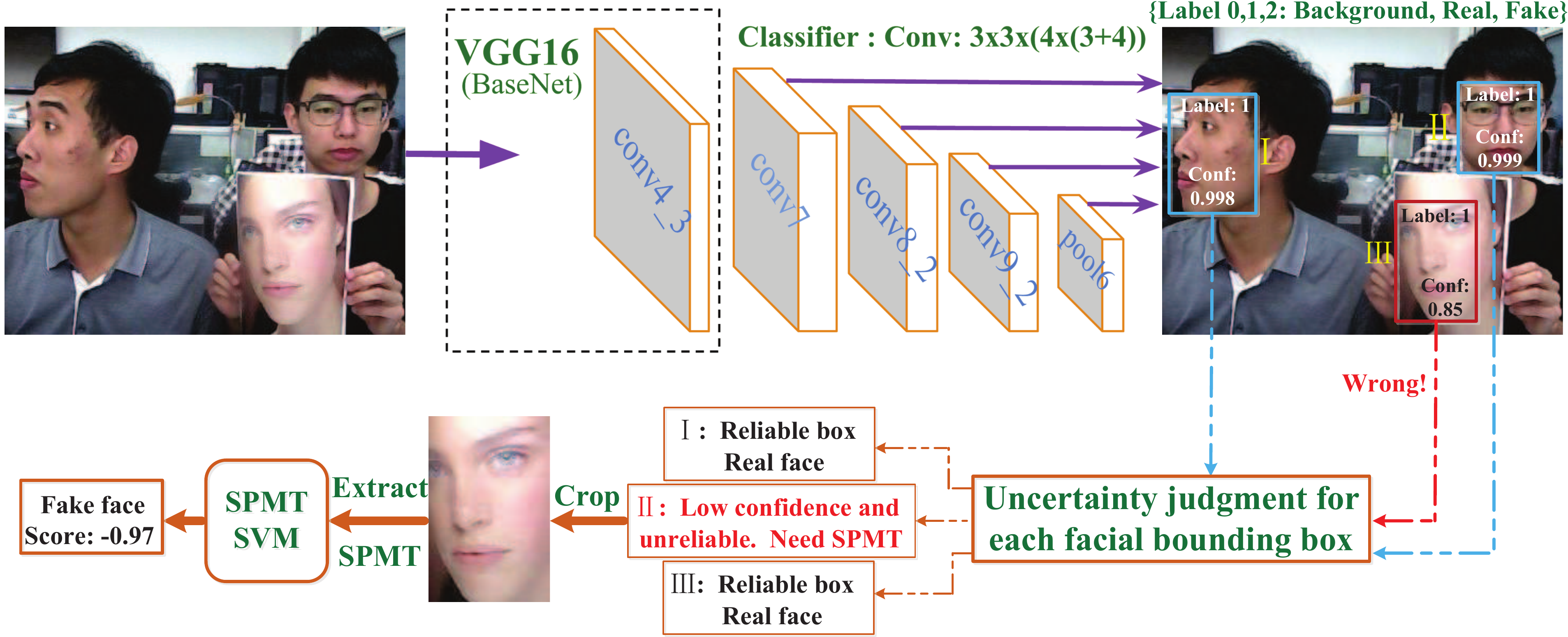}}
\caption{Architecture of the representation combination ``SPMT+SSD''. The image is only an example and not included in our dataset. The lower part of this figure is an illustration of decision-level cascade strategy. SPMT feature is detailed in Fig. \ref{f4}.}
\label{f2}
\end{figure*}

\section{Representation Combination ``SPMT+SSD''}
\label{s3}
\begin{figure*}[tb]
\makebox[\textwidth][c]{\includegraphics[width=1\textwidth]{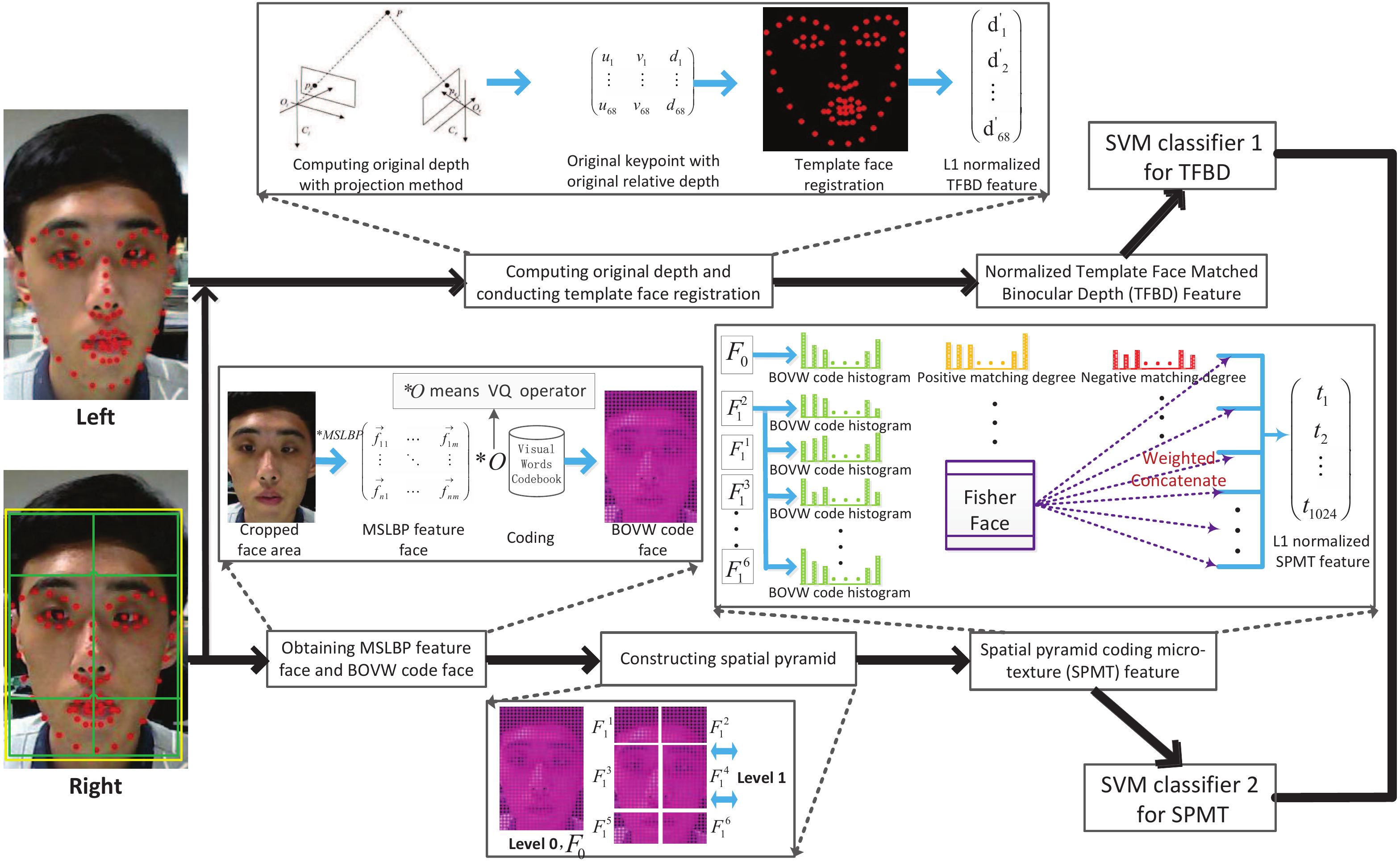}}
\captionsetup{font={footnotesize}}
\caption{Architecture of the representation combination ``SPMT+TFBD''.}
\label{f4}
\end{figure*}

The pipeline is shown in Fig. \ref{f2}. As can be seen, firstly an input image is fed to SSD. SSD can locate all faces in a single image accurately, meanwhile corresponding labels and confidences are also provided. Secondly, the uncertainty judgment is applied to each facial region. For instance, in Fig. \ref{f2}, the liveness judgment for bounding box \uppercase\expandafter{\romannumeral3} is unreliable because its confidence is lower than a previously set threshold. SPMT feature is extracted from an uncertain face (such as bounding box \uppercase\expandafter{\romannumeral3}) and the SVM output decides whether the face is real or not. On the contrary, liveness judgments for certain facial regions (such as bounding box \uppercase\expandafter{\romannumeral1} and \uppercase\expandafter{\romannumeral2}) can be used directly. SPMT descriptor is detailed in Fig. \ref{f4}.
In this section, we will introduce SPMT descriptor, SSD configuration for face presentation attack detection and decision-level cascade strategy.

\subsection{SPMT Feature}

\subsubsection{Basic Setup} First of all, facial regions should be properly cropped. There are two different cropping methods. One method introduced in this section is designed for ``SPMT+SSD'' and another introduced in Section \ref{s4} is designed for ``SPMT+TFBD''. As mentioned in cascade strategy, SPMT should be extracted from uncertain facial regions, which are simply expanded by a ratio of $1.1$ to contain more facial boundaries that are discriminative as demonstrated in \cite{yang2013face}. Then the expanded facial region is cropped, resized to $h_{f}\times w_{f}$ and converted to gray-scale, denoted as $F_{cr}$. Consistent with \cite{yang2013face},   we set $h_{f}$ to $120$ and $w_{f}$ to $100$.

In addition, Local Binary Pattern (LBP) is adopted as the basic micro-texture descriptor and $LBP_{p,r}$ denotes $p$ sampled pixels are on a circular neighborhood with a radius of $r$.

\subsubsection{Fisher Face}
Before illustrating encoding method, we would like to introduce Fisher face, which plays an important role in high-level micro-texture encoding process. Yang \emph{et al} \cite{yang2013face} first introduce Fisher ratio into presentation attack detection. In our work, we will refer to their Fisher criterion analysis, meanwhile making some beneficial improvements.
Inspired by the point that more discriminative parts should be highlighted, Fisher ratio is assigned to each position in $F_{cr}$, to characterize the difference of local micro-texture between real and fake faces. Hence Fisher face $F_{fs}$ owns a same size as $F_{cr}$.

$F_{cr}$ is divided by non-overlapped $10\times 10$ blocks, in order to remain high-frequency information for local difference encoding. Hence, $F_{cr}$ is equivalent to a $h_{f}/10\times w_{f}/10$  block matrix $BM$ and the block matrix preserves the global spatial layout. For each block, three kinds of LBP descriptors \cite{maatta2011face} $\{LBP^{u}_{8,1},LBP^{u}_{8,2},LBP^{u}_{16,2}\}$ are extracted. We concatenate three LBP histograms as a $361$ dimensional vector for each block. Difference of two blocks is calculated by the $\chi^{2}$ distance between two vectors.

One thousand real faces and one thousand fake faces are randomly selected from training set to construct $F_{fs}$. For block $BM_{i,j}$, which denotes the block at position $(i,j)$, we calculate differences between all pairs of $BM_{i,j}$ among selected real faces, then compute their mean and variance: $\mu^{g}_{ij},\sigma^{g}_{ij}$. In a similar way, $\mu^{f}_{ij},\sigma^{f}_{ij}$ for fake faces are obtained. Mean and variance of inter-class difference are denoted as $\mu_{ij},\sigma_{ij}$, calculated by all pairwise distances between two $BM_{i,j}$ in a real and a fake face respectively.
Fisher ratio for block $BM_{i,j}$ is derived by:
\begin{gather}
R_{ij}=\frac{(\mu^{g}_{ij}+\mu^{f}_{ij}-\mu_{ij})^{2}}{\sigma^{g}_{ij}+\sigma^{f}_{ij}-\sigma_{ij}}
\label{e1}
\end{gather}
As can be seen, local neighborhood with small intra-class difference and large inter-class difference is more discriminative.

After Fisher ratios of all $BM_{i,j}$ are calculated, bilinear interpolation is applied to the Fisher ratio matrix, then we obtain a normalized $h_{f}\times w_{f}$ Fisher face.

\subsubsection{Low-Level Descriptor and Mid-level Encoding}

After cropping facial region properly, we conduct low-level and mid-level encoding.
We adopt a MSLBP operator to capture diverse appearance features with various scales, frequencies and orientations, illustrated as $\{LBP_{8,1},LBP_{8,2},LBP_{8,3},LBP_{8,4},LBP_{16,2}\}$. The MSLBP operator is applied to each pixel in $F_{cr}$, obtaining MSLBP feature face $F_{mp}$, which is defined as our low-level texture descriptor. Each pixel in $F_{mp}$ is 48-bit long.

Afterwards, we design a MSLBP codebook $CB$ with a capacity of $N_{cb}$ and use KD-tree generating algorithm to train this codebook. All positive and negative $F_{mp}$ in training set are used, but only a portion of pixels in each $F_{mp}$ are adopted for training. The $k$-th texton in $CB$ is notated as $CB_{k}$, hence codebook is denoted as $\{CB_{k}|1 \leq k \leq N_{cb}\}$. Each texton is $48$-bit long. In our final settings, $N_{cb}=256$.

Then the mid-level texture descriptor called BOVW (Bag of Visual Words) code face $F_{bw}$ is introduced. BOVW coding algorithm in Eq. $(\ref{e2})$ is applied to each pixel in $F_{mp}$, obtaining a $h_{f}\times w_{f}$ BOVW code face:
\begin{gather}
F_{bw}^{(i,j)}=\mathop{\arg\min}_{k}||F_{mp}^{(i,j)}-CB_{k}||^{2}
\label{e2}
\end{gather}
where $1 \leq k \leq N_{cb}$, $F_{bw}^{(i,j)}$ denotes the BOVW code value at position $(i,j)$ in $F_{bw}$, $F_{mp}^{(i,j)}$ denotes the $48$-bit MSLBP feature vector at position $(i,j)$ in $F_{mp}$.

\subsubsection{Spatial Pyramid and the First Part of High-Level Encoding}
Based on the mid-level texture descriptor and pre-trained Fisher face, we conduct high-level encoding based on specifically designed spatial pyramid. Inspired by Spatial Pyramid Matching \cite{lazebnik2006beyond}, a high-level spatial pyramid coding algorithm is proposed.
In the classical spatial pyramid, the grid at level $l$ has $2^l$ pieces along each spatial dimension. However this partitioning method is not appropriate for facial regions. If doing so, facial components will be split chaotically and structure information will be lost. Hence we design a specific spatial pyramid to partition the facial region, preserving the symmetries of facial components.

Level $0$ of spatial pyramid represents the whole facial region. At level $1$, facial region is divided into $3\times 2$ sub-regions. So sizes of sub-regions at level 1 are $\{h_{f}/4\times w_{f}/2, h_{f}/4\times w_{f}/2,h_{f}/2\times w_{f}/2,h_{f}/2\times w_{f}/2,h_{f}/4\times w_{f}/2, h_{f}/4\times w_{f}/2\}$ respectively. From level $2$, partitioning is as same as the traditional method.
Let $F^{s}_{bw,l} ,\, F^{s}_{fs,l}$ denote the $s$-th sub-region at level $l$ in BOVW code face and Fisher face respectively.

The first part of high-level encoding is constructing BOVW histograms for all sub-regions, which are weighted by Fisher face:
\begin{gather}
\begin{array}{l}
\displaystyle BH^{s}_{l}[k]=\frac{1}{|F^{s}_{bw,l}|}\sum_{(i,j)\in{F^{s}_{bw,l}}}F_{fs,l}^{s,(i,j)}\mathbf{I}(F_{bw,l}^{s,(i,j)}=k)
\end{array}
\label{e3}
\end{gather}
where $1 \leq k \leq N_{cb}$, $\mathbf{I}(\cdot)$ is the indicator function, operator $|\cdot|$ counts pixel numbers, $BH^{s}_{l}$ denotes the weighted BOVW histogram of the $s$-th subregion at level $l$.

In our experiments, we construct two-level pyramids on both $F_{bw}$ and $F_{fs}$, hence $7$ regions and $7$ BOVW histograms are obtained in total.

\subsubsection{Class-Specific Face and the Second Part of High-Level Encoding}
Next, we will introduce the second part of high-level encoding for SPMT descriptor, based on proposed class-specific face.
The class-specific face $F_{cs}$ is defined to characterize intra-class similarities. Genuine-specific face $F_{cs,g}$ characterizes the most common BOVW value at each position among genuine faces. Fake-specific face $F_{cs,f}$ can be interpreted in a similar way. For simplicity, we denote all genuine faces and all fake faces in training set as $\Omega_{g}$ and $\Omega_{f}$ respectively. The class-specific face is derived by:
\begin{gather}
F_{cs,\gamma}^{(i,j)}=\mathop{\arg\max}_{k}\sum_{q\in{\Omega_{\gamma}}}\mathbf{I}({F_{bw}^{(i,j)}}^{q}=k)
\label{e4}
\end{gather}
In Eq.  (\ref{e4}), $1 \leq k \leq N_{cb}$, $\gamma=g$ or $f$. ${F_{bw}^{(i,j)}}^{q}$ denotes BOVW code at position $(i,j)$ of the $q$-th sample among $\Omega_{\gamma}$.
The two-level pyramid is also applied on $F_{cs,g}$ and $F_{cs,f}$, obtaining $7$ regions each, denoted as $\{F_{cs,\gamma,l}^{s}\}$.

We also define the positive matching-degree vector and negative matching-degree vector, in order to capture face's similarities with two class-specific faces. Similarities can be measured in all sub-regions weighted by Fisher face. The matching degree vector is derived by:
\begin{gather}
\begin{array}{l}
 \displaystyle \; \; \; \; \; \,
        M^{s}_{\gamma,l}[k]=\left\{
             \begin{array}{lr}
             \displaystyle 1 \qquad   f^{s}_{l}[k]=0 \, \&\& \, c^{s}_{\gamma,l}[k]=0 &  \\
             \displaystyle min(\frac{f^{s}_{l}[k]}{c^{s}_{\gamma,l}[k]},\frac{c^{s}_{\gamma,l}[k]}{f^{s}_{l}[k]}) \quad others
             \end{array}
\right.
\end{array}
\label{e5}
\end{gather}
where $k\in[1,N_{cb}]$, $l$ denotes the pyramid level, $s$ denotes the $s$-th subregion, $\gamma=g$ or $f$ indicates matching with genuine or fake specific face, $f^{s}_{l}[k]$ denotes occurrence frequency of the $k$-th texton in facial region weighted by Fisher ratio: $f^{s}_{l}[k]=\sum_{(i,j)\in{F^{s}_{bw,l}}}F_{fs,l}^{s,(i,j)}\mathbf{I}(F^{s,(i,j)}_{bw,l}=k)$, ${c^{s}_{\gamma,l}[k]}$ denotes the corresponding frequency in class specific face: $c^{s}_{\gamma,l}[k]=\sum_{(i,j)\in{F_{cs,\gamma,l}^{s}}}F_{fs,l}^{s,(i,j)}\mathbf{I}(F_{cs,\gamma,l}^{s,(i,j)}=k)$.

Two matching-degree vectors are constructed for each subregion and then normalized. Finally, we concatenate $7$ BOVW histograms and $14$ matching-degree vectors from all sub-regions as the SPMT descriptor. In our final settings, each SPMT feature vector is $5376$ dimensional then it's reduced to $1024$ dimension by PCA algorithm.

\subsection{SSD for Face Presentation Attack detection}

\subsubsection{Why We Choose SSD}
SSD discretizes the output space of each feature map to conduct position regression for facial region. Those multiple feature maps with varied resolutions and hierarchies naturally provide diverse semantical descriptions to conduct liveness judgement. The pipeline for traditional presentation attack detection includes two stages: firstly detect and crop facial region and then extract features. However the pipeline for SSD based presentation attack detection method is end-to-end.

\subsubsection{Task Configuration}
Label 0 is assigned to anchor boxes that only contain background information. Label 1 and label 2 are assigned to anchor boxes containing a real face and a fake face respectively. Then, SSD converts the face presentation attack detection to a classical detection problem with three categories.

In four datasets that we use, most images only contain one face. For these images, liveness judgment is the label of output bounding box after non-maximum suppression. However for those images containing multiple faces, more than one output bounding boxes exist. Then we should manually check the liveness judgments under this situation.

\subsubsection{Multi-Scale Anchor Boxes}
SSD arranges a set of anchor boxes with several fixed aspect ratios and scales on each predicted layer. The scales of anchor boxes can be calculated by:
\begin{gather}
sc_{i}=sc_{min}+\frac{(i-1)\times(sc_{max}-sc_{min})}{max-1}
\label{e6}
\end{gather}
where $i\in[2,max-1]$, $max$ denotes amount of predicted layers, $sc_{min},sc_{i},sc_{max}$ represent scales of anchor boxes at lowest, the $i$-th and the highest predicted layer respectively.

\subsubsection{Training Samples Selection}
In training set, anchor boxes with label 1 or 2 are ``positive samples'', negative training samples are the anchor boxes with label 0.

To obtain positive training samples, we firstly find the largest overlapped anchor box with each ground truth. Afterwards we regard an anchor box whose overlap ratio with any ground truth is larger than a threshold ($0.5$), as a positive sample. All remaining samples are regarded as potential negative training samples. The hard negative mining strategy is utilized, in which those with higher confidences are selected as negative training samples. The ratio between negative and positive training samples is $3$:$1$ .

\subsubsection{Data Augmentation for Training}

Powerful data augmentation strategy is utilized to enhance shift invariance and scale invariance. For each input image, cropping and expanding operations are used randomly, meanwhile overlap ratio between each cropped or expanded anchor box and corresponding ground truth should be larger than a threshold. Rotation operation is adopted at a certain probability ($0.5$).

\subsection{Decision-Level Cascade Strategy}
We define two kinds of uncertain facial bounding box. For the first kind, an output bounding box $\delta_{1}$ may be judged as a real face with a high confidence, however another bounding box $\delta_{2}$ which has a large overlap ratio with $\delta_{1}$ is judged as a fake face also with a high confidence. For the second kind, the detection confidence of a bounding box $\delta_{3}$  is lower than a certainty threshold $\theta_{c}$ after NMS. We call $\delta_{1},\delta_{2},\delta_{3}$  uncertain bounding boxes and corresponding liveness judgments from SSD can not be used. In this situation, SPMT descriptor should be extracted from facial regions and the outputs from SVM decide whether these faces are real or not.

An certain facial bounding box always has a high confidence after NMS. In this situation, liveness judgment from SSD can be used directly as the final decision because it's reliable.

The second uncertainty metric is adopted and $\theta_{c}$ is tuned on development set ($\theta_{c}=0.92$). Through the cascade strategy, SPMT descriptor and SSD framework are demonstrated complementary, hence the excellent performance is achieved.

\section{Representation Combination ``SPMT+TFBD''}
\label{s4}
The pipeline is shown in Fig. \ref{f4}. Firstly TFBD feature is extracted from a binocular image pair.
Each detected facial landmark is augmented with the third dimension of relative depth, then it is transformed based on template face registration algorithm to match corresponding landmark in template face. After several rounds of iterative optimizations, TFBD descriptor is extracted. At the same time, SPMT feature is extracted from right image. Finally score fusion of two SVM outputs determines the classification result. In this section, we mainly focus on introducing TFBD descriptor. Some supplemental descriptions for SPMT and classification are also provided.

\subsection{TFBD Feature}
\subsubsection{Basic Setup}
Firstly we set up unparallel dual cameras then stereo calibration is conducted. ``Left image'' and ``right image'' are obtained at the same time.
We utilize the regressing local binary features \cite{ren2014face} to locate $N_{P}$ (in this method $N_{P}=68$) facial landmarks and obtain their pixel coordinates in both left and right images. Point based distortion correction algorithm is then applied to each landmark, and finally stereo rectification is conducted.

\subsubsection{Original Landmark Depth}

In the first place, the depth of all facial landmarks should be calculated and some notations are defined as follows.
$R_{c}$ is defined as the rotation matrix and $t_{c}$ represents the translation vector. $M_{r}=\{f_{xr},c_{xr},f_{yr},c_{yr}\}$ and $M_{l}$ are defined as the intrinsic matrices of right and left cameras respectively. $p_{l}=[u_{l},v_{l},1]^{T}$ and $p_{r}=[u_{r},v_{r},1]^{T}$ are homogeneous pixel coordinates of a certain landmark in left and right images respectively. The  coordinate of a certain landmark under right camera's coordinate system is denoted as $p_{rw}=[x_{rw},y_{rw},z_{rw},1]^{T}$. An intermediate matrix $M$ is defined as $M=M_{l}\left[\begin{array}{lcr}R_{c} & t_{c}  \\0^{T} & 1 \end{array}\right]$.

According to the pinhole camera model, the original depth of a certain facial landmark is calculated by:
\begin{align}
  z_{rw}&=\frac{B_{12}b_{2}-B_{22}b_{1}}{\frac{u_{r}-c_{xr}}{f_{xr}}(B_{12}B_{21}-B_{11}B_{22})+(B_{12}B_{23}-B_{22}B_{13})}
  \vspace*{-10pt}
\label{e8}
\end{align}
where $B_{1j}=m_{1j}-m_{3j}u_{l},B_{2j}=m_{2j}-m_{3j}v_{l},b_{1}=m_{34}u_{l}-m_{14},b_{2}=m_{34}v_{l}-m_{24}$ and $m_{ij}$ represents the element at position $(i,j)$ in matrix $M$.

\subsubsection{3-D Abstract Facial Landmark and Template Face}
We define the 3-D abstract landmark based on the original facial landmark, which is very important for TFBD feature and registration operation. Each face can be represented by a set of $N_{P}$ abstract landmarks: $\{p_{j}|p_{j}=[x_{j},y_{j},d_{j}]^{T},1\leq j \leq N_{P}\}$, where the first and the second dimensions are pixel coordinates and the third dimension denotes relative depth.
As illustrated in the next section, abstract landmarks are transformed in each registration round. Original abstract landmark $p_{j}^{1}$ is defined for the first-round registration, where $x_{j}^{1},y_{j}^{1}$ are pixel coordinate of the $j$-th landmark  in right image and $d^{1}_{j}=z_{rw}^{j}-(\sum_{j}z_{rw}^{j})/N_{P}$ ($z_{rw}^{j}$ denotes original depth of the $j$-th facial landmark).

\begin{figure}[htbp]
\captionsetup{font={footnotesize}}
\centering
\includegraphics[width=5in]{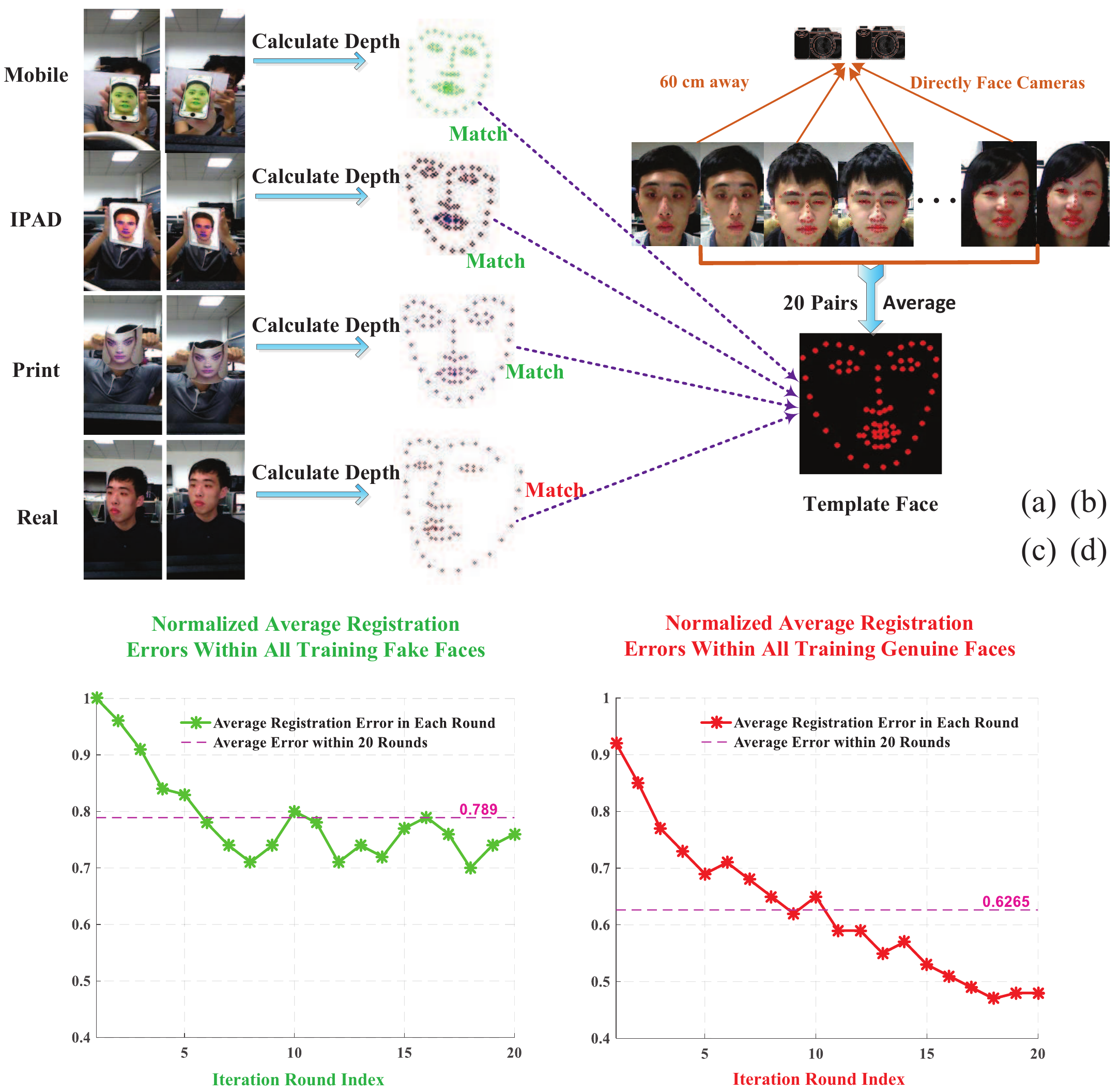}
\caption{(a) Examples of our dataset. (b) Description of template face and registration. (c) and (d) describe iterative registration errors. Generally genuine face can match the template face better with a lower error, which is a powerful cue for classification. All errors have been normalized.}
\label{f5}
%\vspace{-6mm}
\end{figure}

For both real and fake faces, intra-class difference of stereo structure is quite large due to various poses. Consequently a standard real face structure is needed: when the face is similar to this standard structure after a sort of transformation, the face is more likely to be a real face. We name this standard face as template face, which serves as a stereo structure benchmark for face presentation attack detection, as shown in Fig. \ref{f5}.

Template face is obtained before training. We sample $N_{I}$ image pairs from $5$ different people, sitting with varied but moderate distance away from cameras. We set $N_{I}$ to $20$ and distance to \{50cm,60cm,70cm,80cm\}. All cameras should be exactly opposite to people's faces when collecting image pairs for template face. As shown in Fig. \ref{f5}, facial landmarks are located most precisely and depth calculation is most accurate under this situation.

Template face $T$ is represented by a set of standard abstract landmarks : $\{T_{j}|T_{j}=[T_{j}^{x},T_{j}^{y},T_{j}^{d}]^{T},1\leq j \leq N_{P}\}$.
$x_{j}^{1,i}$, $y_{j}^{1,i}$ are notated as pixel coordinate of the $j$-th facial landmark in the $i$-th right picture among $N_{I}$ sampled pairs, and $d_{j}^{1,i}$ represents corresponding original relative depth:
\begin{equation}
  T_{j}^{x}=\frac{1}{N_{I}}\sum_{i=1}^{N_{I}}x_{j}^{1,i},T_{j}^{y}=\frac{1}{N_{I}}\sum_{i=1}^{N_{I}}y_{j}^{1,i},T_{j}^{d}=\frac{1}{N_{I}}\sum_{i=1}^{N_{I}}d_{j}^{1,i}
\label{e9}
\end{equation}

\subsubsection{Template Face Registration Algorithm}
As mentioned above, a sort of transformation should be defined for the original face, to match template face until the ``closest'' degree. This transformation is named as template face registration. As shown in Fig. \ref{f5}, matching error distribution is different among real and fake faces. Hence the TFBD descriptor will utilize this important cue for classification. In a word, ideal registration transformation seeks for optimal parameters to obtain minimal registration error:
\vspace*{-5pt}
\begin{align}
  \mathop{\arg\min}_{s_{F},R_{F},t_{F}}\sum_{j=1}^{N_{P}}{\parallel T_{j}-s_{F}R_{F}\times p_{j}-t_{F}\parallel}^{2}
  \vspace*{-10pt}
\label{e10}
\end{align}
where $s_{F}$ denotes scaling factor, $p_{j}$ denotes the $j$-th abstract landmark, $R_{F}$ and $t_{F}$ denote rotation matrix and translation vector for abstract landmark.

Solving the optimal parameter in Eq. $(\ref{e10})$ directly has poor accuracy. Hence we propose a template face registration algorithm, which is based on unit quaternion absolute orientation method proposed in \cite{horn1987closed}. We modify the algorithm in \cite{horn1987closed} and propose an iterative optimization method. Optimal parameters are estimated by multiple rounds of iterative correction, rather than by single-round calculation.

Single-round registration is defined as:
\begin{equation}
p_{j}^{n+1}={s_{F}^{n}}^{*}{R_{F}^{n}}^{*}\times p_{j}^{n}+{t_{F}^{n}}^{*}
\label{e11}
 \end{equation}
where $n\,(n\geq 1)$ represents the registration round and ${R_{F}^{n}}^{*}$, ${s_{F}^{n}}^{*}$, ${t_{F}^{n}}^{*}$ denote optimal parameters in the $n$-th round. Each abstract landmark $p_{j}^{n}$ is transformed to $p_{j}^{n+1}$ after the $n$-th registration round.

Our point-based iteration method combines the iterative closest point (ICP) algorithm with bootstrapping. We design an abstract landmark pool $Po$, then (${s_{F}^{n}}^{*},{R_{F}^{n}}^{*},{t_{F}^{n}}^{*}$) can be solved based on $Po^{n}$ (the pool used in the $n$-th round). Each $p_{j}^{n+1}$  is calculated according to Eq.  $(\ref{e11})$, meanwhile $N_{P}$ registration errors in this round are obtained: $e_{j}^{n}=\parallel T_{j}-p_{j}^{n+1}\parallel^{2}$. Then all errors are normalized. Afterwards we select $N_{min}$ abstract landmarks with minimal registration error to build $Po^{n+1}$ for the next-round registration.

We assign weight for each landmark in $Po^{n+1}(n \geq 1)$ to highlight the hard examples:
\begin{gather}
     w_{j}^{n+1}=ln\frac{e_{j}^{n}}{1-e_{j}^{n}}
\label{e13}
\end{gather}
where $1 \leq j \leq  N_{min}$, $w_{j}^{n+1}$ denotes weight of the $j$-th landmark in $Po^{n+1}$. Afterwards all weights should be normalized.

The $N_{min}$ landmarks in $Po^{n+1}(n \geq 1)$ has good statistical properties because of small registration errors. Meanwhile according to the idea of bootstrapping, a landmark with a larger error should be assigned with a higher weight for next-round registration. In our final settings, $N_{min}=30$, $Po^{1}$ contains all $N_{P}$ original abstract landmarks and $w_{j}^{1}=1$
 $(1\leq j \leq N_{P})$.

Then for the $n$-th registration round, we use the modified unit quaternion based absolute orientation method to solve the following problem:
\begin{align}
\begin{array}{c}
 \mathop{\arg\min}\limits_{s_{F}^{n},R_{F}^{n},t_{F}^{n}}\sum\limits_{p_{j}^{n}\in{Po^{n}}}{(w_{j}^{n})}^{2}{\parallel T_{j}-s_{F}^{n}R_{F}^{n}\times p_{j}^{n}-t_{F}^{n}\parallel}^{2}
 \end{array}
\label{e14}
 \end{align}

Firstly we define $T^{'}_{j}=w_{j}^{n}T_{j}$, $p_{j}^{'n}=w_{j}^{n}p_{j}^{n}$. Considering $t_{F}^{n}$ is same for each landmark, we define $t_{F}^{'n}=\overline{w_{j}^{n}}t_{F}^{n}$ to approximate the $t_{F}^{n}$, where $\overline{w_{j}^{n}}$ denotes the average weight among $\{w_{j}^{n}\}$. Then Eq.  $(\ref{e14})$ is converted to the standard form as Eq.  $(\ref{e15})$.
\begin{gather}
\begin{array}{l}
 \qquad \quad  \mathop{\arg\min}\limits_{s_{F}^{n},R_{F}^{n},t_{F}^{'n}}\sum\limits_{p_{j}^{n}\in{Po^{n}}}{\parallel err_{j}\parallel}^{2} \vspace*{2pt} \\ \;
  s.t. \quad err_{j}=T^{'}_{j}-s_{F}^{n}R_{F}^{n}\times p_{j}^{'n}-t_{F}^{'n}
 \end{array}
\label{e15}
 \end{gather}

\begin{algorithm}[t]
\caption{Template Face Registration Algorithm}
\hspace*{0.02in} {\bf Input:}
\\ \hspace*{0.1in} $N_{P}$ original abstract landmarks: $\{p_{j}^{1}|1 \leq j \leq  N_{P}\}$
\\ \hspace*{0.1in} Template face $T=\{T_{j}|1\leq j \leq N_{P}\}$
\\ \hspace*{0.1in} Round index $n=1$, $n_{max}=20$\\
\hspace*{0.02in} {\bf Output:}
$N_{P}$ dimensional TFBD descriptor.
\begin{algorithmic}[1]
\State Initialize landmark pool: $Po^{1}\gets \{p_{j}^{1}\}$, $w_{j}^{1}=1$.
\While{$n\leq{n_{max}}$}
¡¡¡¡\State Based on $Po^{n}$, solve $({s_{F}^{n}}^{*}$,${t_{F}^{n}}^{*}$,${R_{F}^{n}}^{*})$ according to the Eq.  $(\ref{e14})$, $(\ref{e15})$, \hspace*{0.16in} $(\ref{e16})$ and $(\ref{e17})$
    \State $p_{j}^{n+1}={s_{F}^{n}}^{*}{R_{F}^{n}}^{*}\times p_{j}^{n}+{t_{F}^{n}}^{*}$, for $1 \leq j \leq  N_{P}$
    \State $e_{j}^{n}=\parallel T_{j}-p_{j}^{n+1}\parallel^{2}$, for $1 \leq j \leq  N_{P}$
    \State Select $N_{min}$ landmarks with minimal errors
    \State Update $Po^{n+1}$ with $N_{min}$ selected landmarks
    \State Update weight $w_{j}^{n+1}$ according to Eq.  $(\ref{e13})$
\EndWhile
\State Concatenate $\{d^{n_{max}+1}_{j}|1\leq j \leq N_{P}\}$ as TFBD descriptor
\end{algorithmic}
\label{a1}
\end{algorithm}

Next we denote $\overline{T^{'}_{j}}$ and $\overline{p^{'n}_{j}}$ as centroids of $T^{'}_{j}$ and $p^{'n}_{j}$ respectively. We use the basic algorithm in \cite{horn1987closed} to solve Eq.  $(\ref{e15})$, then ${s_{F}^{n}}^{*}$ and ${t_{F}^{n}}^{*}$ are obtained:
\begin{gather}
 \ {s_{F}^{n}}^{*}=\frac{\sum\limits_{p_{j}^{n}\in{Po^{n}}} (T^{'}_{j}-\overline{T^{'}_{j}})^{T}\cdot ({R_{F}^{n}}^{*} \times (p_{j}^{'n}-\overline{p^{'n}_{j}}))}{ \sum\limits_{p_{j}^{n}\in{Po^{n}}} {\parallel (p_{j}^{'n}-\overline{p^{'n}_{j}})\parallel}^{2} } \label{e16}\\ \ \
 {t_{F}^{n}}^{*}= (\overline{T^{'}_{j}}-{s_{F}^{n}}^{*}{R_{F}^{n}}^{*}\times \overline{p^{'n}_{j}})/\overline{w_{j}^{n}} \label{e17}
\end{gather}
where `$\cdot$' represents the dot product operation.

Finally, according to \cite{horn1987closed}, rotation matrix is equivalent to an unit quaternion. Let  $\mathring{{q}}^{*}=q_{0}+iq_{x}+yq_{y}+kq_{z}$ denote the optimal unit quaternion. To solve $\mathring{{q}}^{*}$, we define a matrix $Q=\sum_{p_{j}^{n}\in{Po^{n}}}(p_{j}^{'n}-\overline{p^{'n}_{j}})\times(T^{'}_{j}-\overline{T^{'}_{j}})^{T}$
and a $4 \times 4$ matrix $Q^{'}$. As illustrated in Section $4.A$, $p7$, \cite{horn1987closed}, $Q^{'}=\phi(Q)$ where $\phi(\cdot)$ is a matrix transformation operator, meanwhile $\mathring{{q}}^{*}$ equals to the eigenvector corresponding to the maximal eigenvalue of matrix $Q^{'}$. Once $\mathring{{q}}^{*}$ is obtained, ${R_{F}^{n}}^{*}$ can be solved by transforming $\mathring{{q}}^{*}$ according to Section $3.E$, $p6$, \cite{horn1987closed}.

After $n_{max}$ registration rounds, $N_{P}$ transformed relative depth from $N_{P}$ abstract landmarks are concatenated as $N_{P}$ dimensional TFBD descriptor. In our final settings, $n_{max}=20$ and TFBD feature is $68$ dimensional. The overall template face registration algorithm is illustrated in Algorithm \ref{a1}.

\subsection{Supplemental Statements for SPMT and Classification}

Facial region cropping method is different from Sec.\ref{s3}. A face detector based on cascade detection method \cite{viola2001rapid} is employed, meanwhile two eyes are also located. Initially detected facial region can not be used directly because it always contains too much background disturbances. Consistent with \cite{yang2013face}, $D_{eye}$ is defined as the pixel distance between two eyes, the width of cropped facial region $W_{f}$ is $1.6D_{eye}$ and the average of $H_{f}/W_{f}$ ratio is $1.2$. A $H_{f}\times W_{f}$ facial region is then cropped. Finally it's also resized to $h_{f}\times w_{f}$ as in Sec.\ref{s3}.1.

TFBD feature and SPMT feature are individually fed to corresponding nonlinear SVM classifiers. Score fusion of two SVM outputs determines the classification result.

\section{Experiments}
\label{s5}
\subsection{Datasets and Decoding Process}
Four datasets are used in our experiments, including three public datasets: NUAA dataset \cite{tan2010face}, CASIA dataset \cite{zhang2012face} and Replay-Attack dataset \cite{Chingovska_BIOSIG-2012}. As no binocular camera based dataset for face presentation attack detection is publicly available, we construct our own dataset.

\subsubsection{NUAA Dataset} It contains $12641$ still images. Warped photos with different sizes serve as presentation attacks.

\subsubsection{CASIA Dataset} It contains $600$ videos from $50$ subjects in total and covers three kinds of attacks (photo, cut photo, video). For each subject, images of real face and three attacks are captured under three image qualities (low, normal and high qualities). Hence $12$ videos are captured for each subject.

\subsubsection{Replay-Attack Dataset} Replay-Attack dataset contains $50$ subjects and $1300$ videos in total. For each subject, there are two kinds of shooting background (control and adverse), three kinds of attacks (print, digital photo and video), two attacking manners (fixed and hand-holding).

\begin{table}
\footnotesize
\renewcommand{\captionfont}{\small}
\centering
\caption{Compositions of our dataset}
\label{t_our}
\begin{tabularx}{12cm}{|p{3cm}<{\centering}|X<{\centering}|X<{\centering}|X<{\centering}|X<{\centering}|}
\hline
\multirow{3}{*}{Our Dataset} &
\multirow{3}{*}{Genuine} &
\multicolumn{3}{c|}{Spoofing} \\
\cline{3-5}
& & \multirow{2}{*}{Photo} & \multirow{2}{*}{IPAD} & \multirow{2}{*}{Cellphone} \\
& & & &\\
\hline
Image pairs & 6000 & 3000 & 1500 & 1500 \\
\hline
Subjects & 15 & 30 & 15 & 15 \\
\hline
\end{tabularx}
\begin{tablenotes}
\centering
    \item [1] Our dataset is collected using two 640$\times$480 web cameras.
\end{tablenotes}
\end{table}

\begin{table}
\footnotesize
\renewcommand{\captionfont}{\small}
\centering
\caption{Divisions of four datasets after decoding}
\label{t2}
\begin{tabularx}{13cm}{|p{4cm}<{\centering}|X<{\centering}|X<{\centering}|X<{\centering}|X<{\centering}|}
\hline
Dataset & NUAA & CASIA & REPLAY & Ours \\
\hline
Training Subjects & 8 & 20 & 15 & 7+25 \\
\hline
Training Images (Pairs) & 3491 & 45000 & 93000 & 5300 \\
\hline
Test Images (Pairs) & 9150 & 57000 & 124000 & 6700 \\
\hline
Development Set Images & N/A & N/A & 93000 (15) & N/A \\
\hline
\end{tabularx}
\begin{tablenotes}
\centering
    \item [1] 'N/A' means that development set is not divided.
\end{tablenotes}
\end{table}

\subsubsection{Our Dataset} Our dataset is consisted of binocular image pairs, sampled with two fixed and calibrated web cameras with resolution of $640\times480$. The distance between binocular camera is $12$cm. As shown in Table \ref{t_our},
$15$ people are invited, three kinds of presentation attack exist and $60$ fake faces are collected in total. Each person is required to raise head, lower head, rotate face, sit with different positions and varied distance away from cameras. For each fake face, we move it horizontally, vertically, back and front, and rotate it in depth, under different illumination conditions and varied distance. Especially for those printed attacks, we also bend them inward and outward. There are $12000$ image pairs in total.

\subsubsection{Decoding Videos into Frames}
Considering our algorithms are conducted on single images or image pairs, we use each frame in a video. This operation is called ``decoding the video''. Training, test and evaluation are all conducted on still images.
CASIA and REPLAY-ATTACK datasets need to be decoded because they are composed of videos. CASIA dataset contains $102000$ frames and REPLAY-ATTACK contains $310000$ frames in total.

\subsubsection{Dataset Division}  Three public datasets are divided according to \cite{tan2010face,zhang2012face,Chingovska_BIOSIG-2012}. As shown in Table \ref{t2}, `$7+25$' denotes $7$ real faces and $25$ fake faces are selected for training and `$93000 (15)$' means that development set of REPLAY-ATTACK contains $93000$ single images from $15$ subjects.

\subsection{Performance Measures}

To compare with previous works, we adopt Accuracy, Area Under ROC Curve (AUC) \cite{yang2013face}, Equal Error Rate (EER) \cite{yang2013face} that corresponds to the point where false rejection rate (FRR) is equal to false acceptance rate (FAR) in ROC curve, True positive rate (TPR) when FAR is $0.1$ (\cite{wen2015face}) and HTER \cite{Chingovska_BIOSIG-2012}.

Reporting PAD results using only HTER and EER can be biased. Hence to use standardised metrics for evaluation, we follow the ISO standard (ISO/IEC 30107 \cite{timmurphy.org}) and report Attack Presentation Classification Error Rate (\emph{APCER}) and Bona Fide Presentation
Classification Error Rate (\emph{BPCER}).

\subsection{Training and Test Protocols}
For representation combination ``SPMT+SSD'', experiments are conducted on three public datasets. However we can only conduct experiments on our dataset for ``SPMT+TFBD'',  because TFBD descriptor is extracted from a binocular image pair.

We use LIBSVM \cite{chang2011libsvm} to train SVM. For TFBD feature, the category of SVM is ``$\nu$-SVC'' and RBF kernel is used, because TFBD descriptor is only $68$ dimensional and RBF kernel is needed for high dimensional mapping. For SPMT feature, we also use ``$\nu$-SVC''. We use $5$-fold cross validation method to tune hyper-parameters and determine the classification thresholds.

For SSD, we do not tune any hyper-parameters on development set or by K-fold cross validation. All hyper-parameters of network are set empirically, which may not be optimal. However networks with these parameters still achieve nearly perfect performance, revealing the great robustness of SSD for presentation attack detection.

\subsection{Experimental Setup}

For two hand-crafted descriptors, all hyper-parameters in our final settings are described above. For SSD, the input image is resized to $300\times300$ ($500\times500$ for NUAA particularly). We use conv6$\_$2, conv7$\_$2, conv8$\_$2 and conv9$\_$2 for prediction. $sc_{min}$ is set to $0.2$ and $sc_{max}$ is set to $0.9$. We train the network for $20000$ iterations with a learning rate of $10^{-2}$, which is reduced to $10^{-3}$ at $30000$ iterations and $10^{-4}$ at $40000$ iterations. Batch size is $32$. We train and test models on a single NVIDIA Titan-X GPU.

\subsection{Experiments of the Representation Combination ``SPMT+SSD''}
We conduct experiments on three public datasets. For CASIA dataset, we consider seven scenarios including Low Quality (LQ), Methodrate Quality (MQ), High Quality (HQ), Warped Photo (WP), Cut Photo(CP), Video Photo (VP) and Overall test according to the protocols in \cite{zhang2012face}. For REPLAY dataset, we consider seven scenarios including Prints, Mobile, Highdef IPAD (HD), Digital photo (DP), Photo, Video and Overall test according to the protocols in \cite{Chingovska_BIOSIG-2012}.
For each scenario in CASIA and Replay-Attack datasets, we use corresponding subset to train models then conduct evaluation.

\subsubsection{Evaluation of SPMT Descriptor}

\textbf{Results on NUAA dataset}. We make comparisons with four traditional methods, including a re-image theory based method \cite{tan2010face}, a contextual cue based method \cite{komulainen2013context} and two micro-texture based methods \cite{maatta2011face,yang2013face}. MSLBP \cite{maatta2011face} is a state-of-the-art low-level descriptor and Yang's component dependent descriptor \cite{yang2013face} is a state-of-the-art mid-level descriptor. As shown in Table \ref{t1}, our SPMT feature outperforms other traditional descriptors in the literature. We also compare with a convolutional neural network based method \cite{de2017deep}. Our SPMT descriptor is slightly worse than deep texture descriptor from LBPnet \cite{de2017deep}.

\begin{table*}
\footnotesize
\renewcommand{\captionfont}{\small}
\centering
\caption{Performance of the representation combination ``SPMT+SSD'' on NUAA dataset using frame based evaluation metric (Accuracy(\%), EER(\%), AUC)}
\label{t1}
\makebox[\textwidth][c]{
\begin{tabularx}{5.4in}{|p{0.6in}<{\centering}|X<{\centering}|X<{\centering}|X<{\centering}|X<{\centering}|X<{\centering}|X<{\centering}|X<{\centering}|X<{\centering}|}
\hline
&\multicolumn{5}{c|}{Other Descriptors} &\multicolumn{3}{c|}{ \textbf{Ours} }\\
\cline{1-9}
\multirow{2}{*}{Metric} & Tan's & MSLBP & Context & Yang's & LBPnet & \multirow{2}{*}{\textbf{SPMT}} & \multirow{2}{*}{\textbf{SSD}} & \textbf{SPMT}  \\
& \cite{tan2010face}  & \cite{maatta2011face} & \cite{komulainen2013context} & \cite{yang2013face}  & \cite{de2017deep} & & &\textbf{+SSD} \\
\hline

Accuracy  & 88.15  & 92.76 &97.13 & 97.78 & 98.20 & 98.05 & 99.00 & \textbf{99.16}  \\
\hline
AUC  & 0.941 & 0.990 &0.996 & 0.998 & 0.996 & 0.999 &\hspace{-0.3mm}1.000 & \hspace{-0.3mm}\textbf{1.000} \\
\hline
EER  &\hspace{-0.5mm}13.95  &\hspace{-1.4mm}4.84 &\hspace{-1.4mm}2.73 &\hspace{-1.7mm}1.96 & \hspace{-1.2mm}1.80 &\hspace{-1.7mm}1.85 &\hspace{-1.7mm}1.10 & \hspace{-1.5mm}\textbf{0.89}  \\
\hline
\end{tabularx}}
\end{table*}

\textbf{Results on CASIA dataset}. As most state-of-the-art methods don't report \emph{APCER} and \emph{BPCER} for seven scenarios, we only report the performance of our methods, as shown in Table \ref{t4}. The performance for low quality (LQ) is better than higher quality (HQ). This result is expected because fake faces with high-quality usually contain less artifacts. Video presentation attacks are easy to classify due to the inevitable downsize of high-resolution.

\textbf{Results on REPLAY-ATTACK dataset}. In Table \ref{t_re}, an \emph{APCER} of $9.4\%$ and a \emph{BPCER} of $8.3\%$ are obtained on the whole test set, which demonstrates the effectiveness of our spatial pyramid encoding algorithm. However as shown in Table \ref{t5}, our SPMT doesn't achieve state-of-the-art performance due to the local feature's limitations.

\textbf{Discussion}. Our SPMT descriptor focuses on encoding local micro-texture and the results reveal that SPMT outperforms other local descriptors in the term of representation capability.

\begin{table*}
\footnotesize
\renewcommand{\captionfont}{\small}
\centering
\caption{Performance of the representation combination ``SPMT+SSD'' on CASIA dataset using frame based evaluation metric (APCER and BPCER)}
\label{t4}
\makebox[\textwidth][c]{
\begin{tabularx}{5.4in}{|X<{\centering}|X<{\centering}|X<{\centering}|X<{\centering}|X<{\centering}|X<{\centering}|}
\hline
\multicolumn{3}{|c|}{}   &\multicolumn{3}{c|}{ Our Descriptors} \\
\hline
\multirow{2}{*}{Scenario} &Test &\multirow{2}{*}{Metric(\%)}  & \multirow{2}{*}{\textbf{SPMT}} & \multirow{2}{*}{\textbf{SSD}} & \textbf{SPMT+}   \\
&Images &  &  & &\textbf{SSD}   \\
\hline

\multirow{2}{*}{LQ} & 14.1k fake & APCER & 1.42 & 0.85 & \textbf{0.45}   \\
\cline{2-6}
& \hspace{1.4mm}4.6k real &BPCER  & 1.56 & 0.58 &  \textbf{0.34} \\
\hline

\multirow{2}{*}{MQ} & 13.4k fake & APCER & 5.83 & 0.01 & \textbf{0.00}   \\
\cline{2-6}
& \hspace{1.4mm}4.3k real &BPCER  & 5.01 & 0.00 &  \textbf{0.00} \\
\hline

\multirow{2}{*}{HQ} & 15.1k fake & APCER & 7.39 & 0.58 & \textbf{0.35}   \\
\cline{2-6}
& \hspace{1.4mm}4.9k real &BPCER  & 3.55 & 0.27 &  \textbf{0.20} \\
\hline

\multirow{2}{*}{WP} & \hspace{0.4mm}16.9k fake & APCER & 8.11 & 0.75 & \textbf{0.56}   \\
\cline{2-6}
& 13.8k real &BPCER  & 3.02 & 0.19 &  \textbf{0.14} \\
\hline

\multirow{2}{*}{CP} & \hspace{0.4mm}12.8k fake & APCER & 6.83 & 0.67 & \textbf{0.31}   \\
\cline{2-6}
& 13.8k real &BPCER  & 2.52 & 0.12 &  \textbf{0.09} \\
\hline

\multirow{2}{*}{VP} & \hspace{0.4mm}13.4k fake & APCER & 1.50 & 0.11 & \textbf{0.05}   \\
\cline{2-6}
& 13.8k real &BPCER  & 0.45 & 0.04 &  \textbf{0.01} \\
\hline

\multirow{2}{*}{Overall} & \hspace{0.3mm}43.1k fake & APCER & 10.67 & 0.16 & \textbf{0.10}   \\
\cline{2-6}
& 13.8k real &BPCER  & 6.77 & 0.07 &  \textbf{0.04} \\
\hline
\end{tabularx}}
\end{table*}

\begin{table*}
\footnotesize
\renewcommand{\captionfont}{\small}
\centering
\caption{Performance of representation combination ``SPMT+SSD'' on REPLAY-ATTACK dataset using frame based evaluation metric (APCER and BPCER)}
\label{t_re}
\makebox[\textwidth][c]{
\begin{tabularx}{5.4in}{|X<{\centering}|X<{\centering}|X<{\centering}|X<{\centering}|X<{\centering}|X<{\centering}|}
\hline
\multicolumn{3}{|c|}{}   &\multicolumn{3}{c|}{ Our Descriptors} \\
\hline
\multirow{2}{*}{Scenario} &Test &\multirow{2}{*}{Metric(\%)}  & \multirow{2}{*}{\textbf{SPMT}} & \multirow{2}{*}{\textbf{SSD}} & \textbf{SPMT+}   \\
&Images &  &  & &\textbf{SSD}   \\
\hline

\multirow{2}{*}{Print} & \hspace{0.3mm}18.8k fake & APCER & 6.91 & 0.04 & \textbf{0.00}   \\
\cline{2-6}
& 30.0k real &BPCER  & 1.96 & 0.01 &  \textbf{0.00} \\
\hline

\multirow{2}{*}{Mobile} & \hspace{0.5mm}37.6k fake & APCER & 7.50 & 0.16 & \textbf{0.05}   \\
\cline{2-6}
& 30.0k real &BPCER  & 3.64 & 0.04 &  \textbf{0.03} \\
\hline

\multirow{2}{*}{HD} & \hspace{0.5mm}37.6k fake & APCER & 10.99 & 0.01 & \textbf{0.01}   \\
\cline{2-6}
& 30.0k real &BPCER  & 5.81 & 0.00 &  \textbf{0.00} \\
\hline

\multirow{2}{*}{Photo} & \hspace{0.5mm}56.4k fake & APCER & 8.50 & 0.17 & \textbf{0.06}   \\
\cline{2-6}
& 30.0k real &BPCER  & 5.66 & 0.07 &  \textbf{0.04} \\
\hline

\multirow{2}{*}{DP} & \hspace{0.5mm}37.6k fake & APCER & 4.97 & 0.13 & \textbf{0.10}   \\
\cline{2-6}
& 30.0k real &BPCER  & 4.74 & 0.05 &  \textbf{0.03} \\
\hline

\multirow{2}{*}{Video} & \hspace{0.5mm}37.6k fake & APCER & 6.38 & 1.19 & \textbf{1.07}   \\
\cline{2-6}
& 30.0k real &BPCER  & 2.97 & 0.42 &  \textbf{0.27} \\
\hline

\multirow{2}{*}{Overall} & \hspace{0.5mm}94.0k fake & APCER & 9.43 & 0.09 & \textbf{0.04}   \\
\cline{2-6}
& 30.0k real &BPCER  & 8.14 & 0.08 &  \textbf{0.03} \\
\hline
\end{tabularx}}
\end{table*}

\subsubsection{Evaluation of SSD}
$\\$
\hspace*{0.33cm}\textbf{Results}. Images from NUAA dataset are resized to $500\times500$ for training because NUAA training set contains only $3400$ images. Using larger images for training can ease the problem of overfitting.

For CASIA and REPLAY-ATTACK datasets, the size of $300\times300$ is adopted for training because training set is large enough. $300\times300$ model is $2.4$ times faster than $500\times500$ model. As shown in Table \ref{t4} and \ref{t_re}, the \emph{APCER} on total test test is $0.16\%$ for CASIA and $0.09\%$ for REPLAY-ATTACK respectively, which already outperform other methods significantly. But there's still room for improvement, because SSD mainly utilizes global context cues while neglects local features in the facial region.

\textbf{Comparisons with other deep learning based methods}. We individually compare SSD with deep learning based PAD methods (\cite{yang2014learn,menotti2015deep}). As shown in Table \ref{t5}, SSD outperforms other deep networks on CASIA and REPLAY-ATTACK datasets significantly.

\textbf{Discussion}. As can be seen from Table \ref{t1}, SSD doesn't achieve such amazing performance on NUAA dataset, because training set is too small, data diversity is limited and the image quality is quite poor.

\begin{figure}[tbp]
\captionsetup{font={footnotesize}}
\centering
\includegraphics[width=\textwidth]{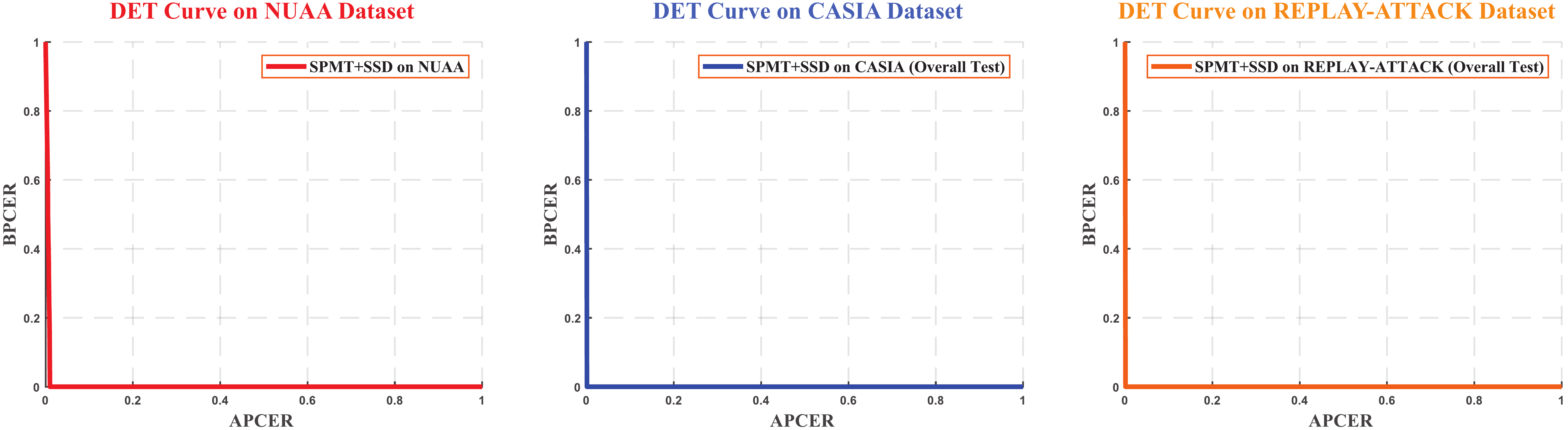}
\caption{DET curves of the representation combination ``SPMT+SSD'' on NUAA, CASIA and REPLAY-ATTACK datasets.}
\label{froc}
\end{figure}

\subsubsection{Evaluation of the combination ``SPMT+SSD''}

\textbf{Results}. As shown in Table \ref{t1}, \ref{t4} and \ref{t_re}, our proposed face PAD method achieves accuracies of more than $99\%$ on all datasets and scenarios. The first representation combination ``SPMT+SSD'' achieves the best performance on NUAA dataset, outperforming the deep texture model LBPnet \cite{de2017deep}. On CASIA and REPLAY-ATTACK datasets, the overall \emph{APCER} and \emph{BPCER} are lower than $0.1\%$, demonstrating the effectiveness of proposed representation combination for 2D face PAD problem. We also present DET curves in Fig. \ref{froc}. As can be seen, performance on single dataset is excellent.

\textbf{Discussion}. Our micro-texture descriptor is highly complementary with deep network SSD, hence local features in the facial region and context cues of the scene can be both utilized. Also the proposed decision-level cascade strategy provides double insurance for face PAD task.

\begin{table}
\footnotesize
\renewcommand{\captionfont}{\small}
\centering
\caption{Comparisons between the ``SPMT+SSD'' and state-of-the-art methods on CASIA/REPLAY-ATTACK benchmarks using frame based evaluation metric}
\label{t5}
\begin{tabularx}{13cm}{|p{3cm}<{\centering}|X<{\centering}|X<{\centering}|X<{\centering}|X<{\centering}|X<{\centering}|}
\hline
\multirow{3}{*}{Method} & \multicolumn{2}{c|}{REPLAY } & \multicolumn{2}{c|}{CASIA} & Speed \\
\cline{2-6}
& HTER  &EER  & EER  & TPR(\%)  &\multirow{2}{*}{fps} \\
& (\%) &(\%)  &(\%) & FAR=0.1 & \\
\hline
LBP+LDA \cite{de2013can} & 19.62 & 18.25  & 21.01 & 75.7 & 5.2 \\
\hline
IQA \cite{Galbally2014Face} & 15.23 & \hspace{-1mm}-  &\hspace{-0.2mm}32.46 & - & -\\
\hline
CDD  \cite{yang2013face} &10.32 & \hspace{-1mm}9.75  & \hspace{-0.5mm}11.85 & 88.8 & 2.5 \\
\hline
\textbf{SPMT} &\hspace{-1mm}9.85 & \hspace{-1mm}9.37  & \hspace{-0.5mm}11.29 & 88.5 & 1.5\\
\hline
Dynamic \cite{de2014face} &\hspace{-1mm}7.65 & \hspace{-1mm}6.76  &\hspace{-0.5mm}10.00 & 89.1 & -\\
\hline
IDA  \cite{wen2015face} &\hspace{-1mm}7.41 & \hspace{-1mm}-  &\hspace{-0.5mm}12.97 & 86.7 &3.8\\
\hline
IQM \cite{Costa2016The} &\hspace{-1mm}5.23 & \hspace{-1mm}- &\hspace{-1.5mm}- & - & -\\
\hline
Person  \cite{yang2015person} &\hspace{-1mm}3.62 & \hspace{-1mm}1.55  &\hspace{-1.8mm}1.63 & - & -\\
\hline
Color \cite{boulkenafet2016face} &\hspace{-1mm}2.81 & \hspace{-1mm}0.42  &\hspace{-1.6mm}2.17 & - & - \\
\hline
CNN \cite{yang2014learn} &\hspace{-1mm}2.75 & \hspace{-1mm}- &\hspace{-1.5mm}6.27 & - & -\\
\hline
SpoofNet \cite{menotti2015deep} &\hspace{-1mm}0.75 & \hspace{-1mm}- &\hspace{-1.7mm}- & - & \hspace{1.2mm}69.0 \\
\hline
\textbf{SSD} &\hspace{-1mm}0.09 & \hspace{-1mm}0.07  &\hspace{-1.6mm}0.08 & \hspace{0.8mm}100.0 & \textbf{120.0} \\
\hline
\textbf{SPMT + SSD} &\hspace{-1mm}\textbf{0.06} & \hspace{-1mm}\textbf{0.04}  &\hspace{-1.6mm}\textbf{0.04} & \hspace{0.8mm}\textbf{100.0} & \hspace{1.2mm}45.5 \\
\hline
\end{tabularx}
\begin{tablenotes}
\centering
    \item [1] -  The value is not provided in corresponding paper.
\end{tablenotes}
\end{table}

\subsubsection{Comparisons with the State-of-the-Art Methods}

NUAA dataset is not considered in this section because few state-of-the-art methods report experimental results on it. As most state-of-the-art methods don't report the standardised metrics as Table \ref{t4}, we use HTER, EER and TPR to compare different PAD methods on CASIA and REPLAY-ATTACK datasets. In Table \ref{t5}, based on frame based evaluation metric, the representation combination ``SPMT+SSD'' for face PAD outperforms other state-of-the-art methods. There are only $50$ wrongly judged images in CASIA and $49$ wrongly judged images in REPLAY-ATTACK dataset.

\subsubsection{Computation Cost Analysis}
The speed of each method reported in Table \ref{t5} is tested on CASIA dataset and the image resolution is $480\times640$. For SPMT, we use a modified cascade detector \cite{viola2004robust} from Matlab Toolbox to detect face and the speed is $0.08$s. Then SPMT descriptor is extracted and the speed is $0.58$s. Hence the total time is $0.66$s per image ($1.5$ fps). For end-to-end SSD, the system runs at $120$ fps without proposals. For ``SPMT+SSD'', SSD is cascaded with SPMT descriptor when uncertain judgements appear. Uncertain judgements are quite rare but they still slow down the whole system to $45.5$ fps in average.

SPMT is tested on a CPU with $32$ GB memory and SSD is tested on a NVIDIA Titan-X GPU. ``SPMT+SSD'' runs on both CPU and GPU. SpoofNet \cite{menotti2015deep} is tested on a NVIDIA TITAN GPU while LBP \cite{de2013can}, CDD  \cite{yang2013face} and IDA  \cite{wen2015face} are tested on a modern CPU.

\subsection{Experiments of the Representation Combination ``SPMT+TFBD''}

\begin{table}
\footnotesize
\renewcommand{\captionfont}{\small}
\centering
\caption{Comparisons of micro-texture descriptors on our dataset}
\label{t6}
\begin{tabularx}{12cm}{|p{2.4cm}<{\centering}|X<{\centering}|X<{\centering}|X<{\centering}|X<{\centering}|X<{\centering}|X<{\centering}|}
\hline
\multirow{2}{*}{Metric} & DOG & Tan's & MSLBP & Context & Yang's  &\textbf{SPMT}  \\
& \cite{zhang2012face} & \cite{tan2010face}  & \cite{maatta2011face} & \cite{komulainen2013context} & \cite{yang2013face}  &\textbf{(Ours)} \\
\hline
Accuracy(\%) & 83.72 & 85.85 & 89.47 & 92.16 & 94.71 & \textbf{94.83}\\
\hline
AUC &0.849 &0.871 &0.917 & 0.958 &0.975 &\textbf{0.978}\\
\hline
EER(\%) &\hspace{-0.5mm}16.13 &\hspace{-0.5mm}14.23 &\hspace{-0.5mm}12.91 &\hspace{-1.5mm}8.47 &\hspace{-1.5mm}6.15 &\hspace{-1.5mm}\textbf{6.02} \\
\hline
\end{tabularx}
\end{table}

\subsubsection{Performance of SPMT on Our Dataset}
The results are shown in Table \ref{t6}. The performance of all descriptors is worse than their performance on NUAA dataset, but our SPMT still outperforms others. It's not surprising because our dataset is more challenging than NUAA because of different head poses and distance during collection. When face is rotated or far away from camera, it is difficult to locate facial region accurately and more background disturbances are included. In order to reduce the sensitivity, binocular depth feature should also be utilized.

\subsubsection{Experiments of TFBD Feature and the Combination ``SPMT+TFBD''}
Results shown in Table \ref{t7} reveal that original depth feature is somewhat discriminative, however when face is increasingly far from cameras, relative depth difference between different landmarks is reduced. Performance also becomes worse when face in front of camera is rotated due to the inaccurate landmark locations. After matched with template face, normalized binocular depth feature can reflect the face stereo structure. The obvious decline in \emph{APCER} and \emph{BPCER} proves the effectiveness of TFBD feature. But TFBD feature is sensitive to some presentation attacks which are very similar to real faces. Hence SPMT feature is introduced and effectiveness is proved by the \emph{APCER} of $8.11\%$. However, limitation also exists when there are too much background disturbances or image quality is high. Hence we combine TFBD descriptor with SPMT descriptor eventually. As can be seen, \emph{APCER} finally drops to $5.95\%$ and EER drops to $3.53$. Score fusion ratio is $1:1$.

\begin{table}
\footnotesize
\renewcommand{\captionfont}{\small}
\centering
\caption{Performance of the combination ``SPMT+TFBD'' on our dataset}
\label{t7}
\begin{tabularx}{12cm}{|p{2.4cm}<{\centering}|X<{\centering}|X<{\centering}|X<{\centering}|X<{\centering}|}
\hline
Operator & Original & TFBD & SPMT & \textbf{SPMT+}  \\
    &depth &feature &feature &\textbf{TFBD}   \\
\hline
APCER(\%) & 22.21 & 10.28 & 8.11 & \textbf{5.95} \\
\hline
BPCER(\%)& 7.12 & 4.18 & 2.34 &\textbf{2.05} \\
\hline
EER(\%) & 14.35 & 8.12 & 6.02 &\textbf{3.53}  \\
\hline
\end{tabularx}
\end{table}

\subsection{Aggregate Dataset Experiments}

In prior work \cite{boulkenafet2016face,de2014face,pinto2015face}, when conducting cross-dataset experiments, the model is trained on one dataset and then test on other datasets. In a degree, it's not meaningful because the generalization ability should be proved in real applications under various situations, rather than measured on a single dataset. Both training and test data should be diverse enough to mimic real scenarios. In addition, convolutional neural networks own powerful fitting abilities. Advantages of deep networks will be wasted if only one dataset is adopted for training.

Hence to demonstrate the generalization ability of our proposed representation combination ``SPMT+SSD'' for face PAD, we conduct aggregate dataset experiments, as shown in Table \ref{t9}. SPMT, SSD and ``SPMT+SSD'' are trained on three public datasets, which are then used for evaluation. We randomly select $10000$ real faces and $10000$ fake faces from three datasets to train aggregate dataset models. Aggregate dataset performance of SSD and the combination ``SPMT+SSD'' on CASIA dataset is worse than single dataset performance (Table \ref{t4}). It is acceptable because SSD's fitting ability is too powerful for a single dataset. Excellent aggregate dataset results on three benchmarks prove that any feature representation under any scenario can be learned by cascading deep network with our local descriptor SPMT.

\begin{table}
\footnotesize
\renewcommand{\captionfont}{\small}
\centering
\caption{Aggregate dataset experiments}
\label{t9}
\begin{tabularx}{13.5cm}{|X<{\centering}|X<{\centering}|p{2.5cm}<{\centering}|X<{\centering}|X<{\centering}|X<{\centering}|}
\hline
Train & Test & Method & HTER(\%) & APCER(\%) & BPCER(\%)\\
\hline
& \multirow{3}{*}{NUAA} & SPMT  & 9.43 & 11.61 & 5.45\\
\cline{3-6}
 & & SSD  & 0.78 & 1.10 & 0.51\\
 \cline{3-6}
 On All&  & SPMT+SSD  & \textbf{0.72} & \textbf{1.04} & \textbf{0.43}\\
\cline{2-6}
Three& \multirow{3}{*}{REPLAY} & SPMT  & 14.27 & 13.08 & 9.35\\
\cline{3-6}
  Public & & SSD  & 0.07 & 0.08 & 0.05\\
 \cline{3-6}
 Datasets &  & SPMT+SSD  & \textbf{0.05} & \textbf{0.04} & \textbf{0.04}\\
\cline{2-6}
  & \multirow{3}{*}{CASIA} & SPMT  & 15.36 & 15.10 & 12.35 \\
\cline{3-6}
 &   & SSD  & 0.28 & 0.35 & 0.29\\
\cline{3-6}
& & SPMT+SSD  & \textbf{0.25} & \textbf{0.30} & \textbf{0.26}\\
\hline
\end{tabularx}
\end{table}

\section{Conclusion and Future Work}
\label{s6}

In today's biometric authentication systems, diverse threats of presentation attacks are increasing.
In order to obtain robust solutions for face PAD, two kinds of discriminative representation combinations are proposed in this paper. The first combination incorporates local appearance features, along with global context cues from deep networks. The complementarity between SPMT descriptor and SSD framework, as well as the proposed decision-level cascade strategy, make the combination very effective for preventing 2D presentation attacks. Excellent experimental results on single-dataset and aggregate-dataset, especially the \emph{APCER} which is lower than $0.1\%$, demonstrate its effectiveness and generalization ability. The second combination cooperates binocular depth information with appearance features. The proposed template face registration method can effectively characterize the difference of stereo structures between real faces and presentation attacks. After incorporating with multi-scale texture features, the sensitivity to stereo structures of presentation attacks is alleviated.
As only a binocular camera is needed, the PAD system with TFBD and SPMT descriptors is an effective and cost-efficient alternative in real face recognition applications.

In future work, other advanced convolutional neural networks will be investigated for face PAD algorithms. In addition, we also consider to incorporate TFBD and SPMT descriptors into deep networks for joint training, to obtain the deep fused representations from unified model. It will be a future direction to achieve a detection system for sophisticated counterfeits.

\clearpage

\bibliographystyle{elsarticle-num}
\bibliography{ref}

\end{document}